Extended Abstract

# ECA-LP / ECA-RuleML: A Homogeneous Event-Condition-Action Logic Programming Language

Adrian Paschke

Internet-based Information Systems, Technische Universität München
Adrian.Paschke@gmx.de

## Abstract

Event-driven reactive functionalities are an urgent need in nowadays distributed service-oriented applications and (Semantic) Web-based environments. An important problem to be addressed is how to correctly and efficiently capture and process the event-based behavioral, reactive logic represented as ECA rules in combination with other conditional decision logic which is represented as derivation rules. In this paper we elaborate on a homogeneous integration approach which combines derivation rules, reaction rules (ECA rules) and other rule types such as integrity constraint into the general framework of logic programming. The developed ECA-LP language provides expressive features such as ID-based updates with support for external and self-updates of the intensional and extensional knowledge, transactions including integrity testing and an event algebra to define and process complex events and actions based on a novel interval-based Event Calculus variant.

## 1. Motivation

Recently, there has been an increased interest in industry and academia in event-driven mechanisms and a strong demand for (complex) event- resp. action processing functionalities comes from the web community, in particular in the area of Semantic Web and Rule Markup Languages (e.g. RuleML or RIF). Active databases in their attempt to support automatic triggering of actions in response to events have intensively explored and developed the **ECA paradigm** and **event algebras** to compute complex events. A different approach to event definitions which has for the most part proceeded separately has its origin in the area of knowledge representation (KR) and logic programming (LP). In these **KR event/action logics** the focus is on the formalization of actions/event axioms and on the inferences that can be made from the happened or planned events/actions. This has led to different views and terminologies on event definitions and event/action processing in both domains. In this paper we present an approach which attempts to combine both views and design a **homogeneous Event-Condition-Action Logic Programming language** (**ECA-LP**) [1] which supports, e.g., complex event- and action processing, formalization of ECA rules in combination with derivation rules and transactional ID-based knowledge updates which dynamically change the intensional and extensional knowledge base (KB). We further show how this approach fits into RuleML[1], a standardization initiative for describing different rule types on the (Semantic) Web and introduce the ECA Rule Markup Language (**ECA-RuleML**) [2] as a XML serialization syntax.

## 2. ECA-LP: A Homogeneous ECA LP Language

*ECA-LP* [1, 2] is a reactive logic programming language developed in the ContractLog KR[2] [3, 4] where reactive rules (ECA rules), derivation rules, goals (queries) and facts are represented together in a **homogeneous knowledge base** (KB): $KB = <R, E, F, I>$, where $R$ is the set of derivation rules, $E$ the set of ECA rules, $F$ the set of facts and $I$ the set of integrity constraints. The KB might be changed over time, i.e. the intensional and extensional knowledge might be updated dynamically during runtime incorporating self-updates triggered by ECA rules. The approach is based on the logic programming paradigm where the KB is an extended logic program (ELP) with finite functions, variables, extended with (non-monotonic) default negation (negation-as-finite-failure) and explicit negation, a typed logic supporting external Java types and procedural attachments using Java objects/methods in rule derivations [5] and Semantic Web types defined in external ontologies (RDFS/OWL) [6, 7]. In the following we use the standard LP notation with an ISO Prolog related scripting syntax called Prova [8] and we assume that the reader is familiar with logic programming techniques [9].

### Syntax of ECA-LP

A reactive rule in ECA-LP is formalized as an extended ECA rule, represented in the KB as a 6-ary fact with the restricted predicate name "*eca*": ***eca(T,E,C,A,P,EL),*** where $T$ (time), $E$ (event), $C$ (condition), $A$ (action), $P$ (post condition), $EL$(se) are complex terms/functions [10]. The complex terms are meta-interpreted by the ECA interpreter as (sub-) goals on further derivation rules in the KB which are used to implement the functionality of each part of an ECA rule. Hence, the full expressiveness of derivation rules in extended LPs with logical connectives, variables, finite functions, (non-monotonic) default and explicit negation as well as linear sequential operators such as "cuts" and procedural attachments can be used to describe complex behavioural reaction logic, whereas the ECA rules' syntax stays compact and the implemented derivation rules can be reused several times in different ECA rules. The *time part* (T) of an ECA rule defines a pre-condition (an explicitly stated temporal event) which specifies a specific point in time at which the ECA rule should be processed by the ECA processor, either absolutely (e.g., "at 1 o'clock on the 1st of May 2006), relatively (e.g., 1 minute after event X was detected) or periodically ("e.g., "every 10 seconds"). The *post-condition* (P) is evaluated after the action has been executed. It might be used to prevent backtracking from different variable bindings via cuts or it might be used to apply post-conditional integrity and verification/validation tests in order to safeguard transactional knowledge updates in ECA rules. The *else part* (EL) defines an alternative action which is executed in case the ECA rule could not be completely executed. This is in particular useful to specify default (re-)actions or trigger exceptional failure handling policies. ECA parts might be blank, i.e. always satisfied, stated with "_", e.g., eca(time(), event(...), _, action(...),_,_). Blank parts might be completely omitted leading to specific types of reactive rules, e.g. standard ECA rules (ECA: eca(event(),condition(),action()) ), production style rules (CA: eca(condition(),action()) ), extended ECA rules with post condition (ECAP: eca(event(),condition(),action(), postcondition()) ).

### Forward-Directed Operational Semantics of ECA-LP

ECA rules, which are homogeneously represented in the KB together with derivation rules, facts and top queries are meta-interpreted by an additional **ECA interpreter**. [10] The interpreter implements the forward-directed operational semantics of the ECA paradigm. An ECA rule (fact) is interpreted as an ordered query (top goal) on the KB, where each of its six literals (positive or negative complex terms/functions) denotes a subgoal. The ECA interpreter provides a general **Wrapper interface** which can be specialized to a particular query API of an arbitrary backward-reasoning inference engine[3], i.e. the ECA meta interpreter is used as a general add-on attached to a LP system extending it with reasoning and processing features for ECA rules. The task of processing an ECA rule by querying the respective derivation rules using the defined complex terms in an ECA rule as queries on the KB is solved by a **Daemon** (implemented within the ECA interpreter). The daemon is a kind of adapter that frequently issues queries on the ECA rules in order to simulate the active behavior in passive goal-driven LP systems: (1) First it queries (repeatedly - in order to capture updates to reactive rules) the KB and derives all ECA rules represented in the KB by the universal query "*eca(T,E,C,A,P,EL)?*", (2) adds the derived ECA rules to its internal active KB, which is a kind of volatile storage for reactive rules and temporal event data, and (3) finally processes the ECA rules

---

[1] http://www.ruleml.org
[2] http://ibis.in.tum.de/staff/paschke/rbsla/index.htm or https://sourceforge.net/projects/rbsla

[3] or production rule system, where the ECA rule terms are used as constraints on the fired conclusions

in parallel using a thread pool. The complex terms defined within an ECA rule are used as queries (top goals) on the globally defined derivation rules in the KB, which implement the functionality of the ECA rule. Note that, to incorporate external procedural logic and functionalities (systems) expressive procedural attachments are used. Figure 1 illustrates the operational semantics of the ECA interpreter.

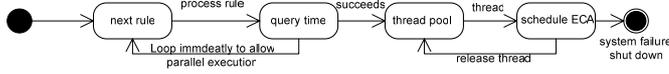

Figure 1: **Concurrent execution of ECA rule using threads**

In particular, this means that the LP inference engine is used to compute variable bindings (applying a typed unification with procedural attachments (see [5] and [6, 7]) and logical connectives including automated rule chaining based on resolution with unification and backtracking to build-up complex behavioral ECA logic in terms of derivation rules. The ECA interpreted itself implements variable unification with backtracking between the ECA parts (subgoals of the top goal) as well as direct support for Java-based procedural attachments exploiting Java reflection to dynamically instantiate Java objects and call their methods and attributes at runtime. The homogeneous combination of ECA rules and derivation rules within the same representation language paves the way to (re-)use various other useful logical formalisms, such as defeasible rules with rule priorities for conflict resolution, event/action logics for temporal event calculations and complex event/action reasoning and it relates ECA rules to other rule types such as integrity constraints or normative rules. It has been shown that such formalisms can be adequately represented as meta programs based on derivation rules (see the expressive ContractLog KR for more information [3, 4]).

In a nutshell, the procedural semantics of reactive rules in ECA-LPs is implemented by the ECA interpreter, which meta-interprets the ECA functions $ECA_i \in KB$ as top queries consisting of a conjunction of literals (the positive, explicit negated or default negated subgoals formalized as complex terms or the Boolean-valued procedural attachments on external Java methods). The else part defines an exception to the normal execution sequence which ends with the standard action / post condition and hence is interpreted as a disjunction: $ECA_i ? = T \wedge E \wedge ((C \wedge A \wedge P) \vee EL)$. The interpreter implements an ordered computation rule (selection function) on the subgoals, selecting the most left goal first, in order to implement the forward-directed operational semantics of the ECA paradigm. Proof-theoretically it applies the ECA subgoals one after the other on the KB using the inference rules of the underlying backward-reasoning inference engine to deductively prove the syntactic derivability from the clauses in the KB, i.e. derive $\Leftarrow ECA_i$ from the facts and derivation rules which are actually in the KB (i.e. from the actual knowledge state): $KB_P \vdash ECA_i$. The computed (ground) substitutions θ of the variables for each subgoal in $ECA_i$ are unified by the ECA interpreter with their variable variants in the other ECA subgoals of the top ECA query. The interpreter also implements the common backtracking mechanism to backtrack from different variable bindings. It is worth noting, that the implementation of the ECA interpreter is designed to be general and to be applicable to different derivation mechanisms such as variants of linear SLDNF-resolution or other non-linear approaches such as SLG-resolution. However, depending on the configuration of the ECA interpreter for a particular logic class (definite, stratified, normal, extended, disjunctive LP) and depending on the procedural semantics of the underlying inference engine, e.g. standard SLDNF with negation-as-finite-failure rule (Naf), query answering might become undecidable (due to infinite functions), non-terminating (due to loops) or floundering (due to free variables in Naf tests). This amounts for configuring the ECA interpreter with special safety conditions such as Datalog restriction or "allowedness" restriction for variables in default negated literals. Our particular ECA-LP reference implementation is based on the backward-reasoning rule engine Prova [8] which provides linear SLDNF resolution, Java object-valued terms and a Java-typed unification with support for Java based procedural attachments [5]. We have extended the Prova inference engine in the ContractLog KR with different selectable and configurable variants of SLDNF implementing e.g. *goal memorization, loop prevention, selection functions or partial reasoning* in order to overcome well-known problems of standard SLDNF resolution and compute extended WFS [11]. In contrast to non-linear tabling approaches such as SLG-resolution we preserve the linearity property while providing goal memorization and hence still enable efficient memory structures and strictly sequential operators to be used.

**Declarative Semantics of ECA-LP**

The declarative semantics of ECA rules in ECA-LPs is directly derived from the semantics of the underlying inference system. ECA rules are defined globally and unordered in the KB. Model-theoretically the semantics is a subset of the 3-valued (Herbrand) models of the underlying rule language: $SEM(ECA-LP) \subseteq MOD_{3-val}^{Herb}(ECA-LP)$ with a skeptical entailment relation $\models$, where $SEM^{skept}(ECA-LP) := \bigcap_{M \in SEM(ECA-LP)} \{M \models L : literal$ is the set of all atoms that are true in all models of $SEM(ECA-LP)$ w.r.t. to updates of the KB (see later). Note that this general definition also covers 2-valued semantics such as Clark's completion semantics, as a declarative counterpart of the procedural "Finite-Failure SLDNF", because from a classical viewpoint the 2-valued models are special 3-valued ones, i.e. any semantics based on a two-valued theory $\Sigma$ can be considered as a three-valued interpretation $I=(T;F)$ with $T=\{A: \text{ a ground atom with } \Sigma \models A\}$ and $F=\{A: \text{ a ground atom with } \Sigma \models neg A\}$. Since in the ContractLog reference implementation of ECA-LP the default semantics is the **well-founded semantics (WFS)** resp. extended WFS (for extended LPs with default and explicit negation) by definition *SEM(ECA-LP)* only contains a single model for the actual knowledge state of the KB: $SEM_{WFS}(ECA-LP) = \{M\}$. Unlike other 3-valued semantics such as Fitting's semantics which are based on Kleene's strong 3-valued logic, WFS [12] is an extension of the perfect model semantics[4]. It is worth noting that WFS can be considered as an approximation of stable models for a program $P$ in the sense that it coincides for a broad class of weakly stratified LPs with STABLE and classifies the same atoms as true or false as STABLE. Hence, roughly speaking, an ECA rule *eca* in an ECA-LP $P$ (or more precisely in the actual knowledge state $P_i$) is satisfied in a finite interpretation $I_{WFS}$, iff there exists at least one model for the sentence formed by the ECA top goal: $P_i \models eca$ and consequently $P_i \vdash eca \Leftrightarrow P_i \models eca$.

While this definition of soundness and completeness for ECA-LPs is sufficient in one particular knowledge state, ECA rules typically deal with knowledge updates which can not only be used to change the extensional knowledge (facts) but also the intensional derivation and behavioral ECA rules, hence leading to a new knowledge state. Such update actions can then be used as input events in further ECA rules leading to update sequences defined in terms of active rules. Obviously, updates of the intensional KB (a.k.a. dynamic LPs) might lead to confluence and logical conflicts (e.g. conflicting positive vs. negative conclusions), which are not resolved by the standard semantics of normal reps. extended LPs. Hence, this amounts for expressive update primitives with a declarative semantics for dynamically changed LPs. We first define semantics for expressive ID-based updates and unitized KB structures, where clauses are labeled by an ID and bundled to clause sets (modules) having a module ID. The core procedural functionalities for clause sets with labels (properties) have been outlined by the original Mandarax system[5] and further extended by Prova [5] which supports dynamic imports of external LP scripts via their URL, with the URL being the module ID of the fresh module (clause set) in the KB. In the ContractLog KR we have further extended this approach with dynamic, transactional ID-based meta functions to add and remove arbi-trary knowledge as modules to the KB using the module ID for partial reasoning on the clause sets, remove modules from the KB via their ID, rollback transactional updates or prioritize conflicting rules an modules (e.g., *overrides(module id1,module id2)*). For instance, consider the following examples using some of the meta functions implemented in the ContractLog KR:

```
add("./examples/test/test.prova")        % add an externally defined module / LP script
add(id1,"r(1):-f(1). f(1).")             % add rule "r(1):-f(1)." and fact "f(1)."
add(id3,"r(_0):-f(_0), g(_0). f(_0). g(_1).",[1,2])   % update with variables
remove(id1)                              % remove all updates with id
remove("./examples/test/test.prova")     % remove external update

transaction(remove(...))    % transactional remove with internal integrity test on all constraints
partial(append([],[xxx],Out),"ContractLog/list.prova")  % apply append on the module list.prova
```

---

[4] hence, e.g. assigns false to a tautology such as $p \Leftarrow p$
[5] http://sourceforge.net/projects/mandarax

Formally, we define a **positive (add) resp. negative (remove) ID-based update** to a program $P$ as a finite set $U_{oid}^{pos/neg} := \{rule^N : H \leftarrow B, fact^M : A \leftarrow\}$, where $N=0,..,n$, $M=0,..,m$ and *oid* denotes the label of the update, i.e. the unique object id with which it is managed in the unitized KB. Applying $U_{oid}^{pos}$ resp. $U_{oid}^{neg}$ to $P$ leads to the extended knowledge state $P^+ = P \cup U_{oid}^{pos}$ resp. reduced state $P^- = P \setminus U_{oid}^{neg}$. Applying arbitrary sequences of positive and negative updates leads to a sequence of program states $P_0,..,P_k$ where each state $P_i$ is either $P_i = P_{i-1} \cup U_{oid\ i}^{pos}$ or $P_i = P_{i-1} \setminus U_{oid\ i}^{neg}$. In other words, a program $P_i$, i.e. the set of all clauses in the KB at a particular knowledge state $i$, is decomposable in the previous knowledge state plus/minus an update, whereas the previous state consists of the state $i$-2 plus/minus an update and so on. Hence, each particular knowledge state can be decomposed in the initial state $P_0$ and a sequence of updates. We define $P_0 = \emptyset \cup U_0^{pos}$ and $U_1^{pos} = \{P : \text{the set of rules and facts defined in the initial LP } P\}$, i.e. loading the initial LP from an external ContractLog/Prova script denotes the first update.

Based on the semantics of updates we define the semantics of ECA-LP with (self-)update actions which have an effect on the knowledge state as a **sequence of transitions** $<P,E,U> \rightarrow <P',U,U'> \rightarrow <P'',U',U''> \rightarrow .. \rightarrow <P^{n+1},U^n,A>$, where $E$ is an initiating event which triggers the update action $U$ of an ECA rule (in case the ECA condition is satisfied) and transits the initial knowledge state $P$ into $P'$. The update action $U$ might be a further sub-event in another ECA rule(s) (active rule) which triggers another update, leading to a sequence of transitions which ends with a terminating action $A$. $A$ might be e.g. an update action which is not an input event for further updates or an action without an internal effect on the knowledge state, e.g. an external notification action.

To overcome arising conflicts, preserve integrity of the KB in each state and ensure a unique declarative outcome of update sequences (active rules) or complex updates (see complex events/actions in next section) we extend the semantics of updates **to transactional updates** which are safeguarded by **integrity constraints (ICs)** or **test cases (TCs)**. ICs and TCs are used to verify and validate the actual or any hypothetically future knowledge state. [13-16] A transactional update is an update, possibly consisting of several atomic updates, i.e. a sequence of transitions, which must be executed completely or not at all. In case a transactional update fails, i.e. it is only partially executed or violates integrity, it will be rolled back otherwise it will be committed. We define a transactional update as $U_{oid}^{trans} = U_{oid_1}^{pos/neg},..,U_{oid_n}^{pos/neg} \& C_1,..C_m$, where $C_i$ is the set of Ics (or TCs) which are used to test integrity and verify and validate the intended models of the updated clauses/modules. In short, ICs are defined as constraints on the set of possible models and describe the model(s) which should be considered as strictly conflicting. We have implemented a meta program within ContractLog which is used to test (meta interpret) integrity constraints:

1. **testIntegrity()** iterates over all integrity constraints in the knowledge base and tests them based on the actual facts and rules in the knowledge base.
2. **testIntegrity(Literal)** tests the integrity of the literal, i.e. it makes a test of the (hypothetically) added/removed literal, which might be a fact or the head of a rule.

Example:
```
neg(p(x)).                      %fact
integrity(xor(p(x), neg(p(x)))).  % mutual exclusive integrity constraint
testIntegrity(p(x))?            %test integrity
```

The example defines an explicitly negated predicate $p(x)$ and an IC, which states that $p(x)$ and the negation of $p(x)$ are mutual exclusive. Accordingly, the hypothetical test *testIntegrity(p(x))* fails. TCs generalize the concept of ICs.

Model-theoretically the truth of an IC in a finite interpretation $I$ is determined by running the constraint definition as a goal $G_{IC}$ (proof-theoretically, by meta interpretation, as a test query) on the clauses in $P$ or more precisely on the actual state $P_i$. If the $G_{IC}$ is satisfied, i.e. there exists at least one model for $G_{IC}$: $P_i \models G_{IC}$, the IC is violated and $P_i$ is proven to be in an inconsistent state: IC is violated resp. $P_i$ violates integrity iff for any interpretation $I$: $I \models P_i \rightarrow I \models G_{IC}$. For more information on the different types of ICs and the meta implementations in the ContractLog KR see [3, 14, 15]. The ICs are used to test the hypothetically updated program state $P_{i+1}^{hypo} = P_i \cup U_{oid}^{trans\ pos}$ resp. $P_{i+1}^{hypo} = P_i \setminus U_{oid}^{trans\ neg}$ and rollback the (possibly partial executed) transactional update to the state $P_{i+1} = P_i$ or commit the update $P_{i+1} = P_{i+1}^{hypo}$ in case the update preserves integrity w.r.t. to the defined ICs. Hence, if there exists a satisfied IC in the hypothetical/pending state $P_{i+1}^{hypo}$ the transactional update is rolled back:

$P_{i+1} = P_i \cup U_{oid}^{trans\ pos} \Rightarrow P_{i+1}^{hypo} \setminus U_{oid}^{trans\ pos}$ iff exists $P_{i+1} \models IC_j, j=1,..,n$

$P_{i+1} = P_i \setminus U_{oid}^{trans\ neg} \Rightarrow P_{i+1}^{hypo} \cup U_{oid}^{trans\ neg}$ iff exists $P_{i+1} \models IC_j, j=1,..,n$

A commit is defined as $P_{i+1}^{hypo} \Rightarrow P_{i+1}$.

Note, that the procedural semantics for transactional updates is based on the **labelled logic** implemented in ContractLog/Prova [5, 8, 14], where rules and rule sets (modules) are managed in the KB as labelled clause sets having an unique oid (object identifier). Updates are new modules (or changes in existing modules) which are added to the KB, e.g. by importing new rule sets from external scripts. The modules are managed and removed from the KB by their oids. Additional local ICs resp. TCs can be explicitly specified in the ECA post condition part and the ContractLog KR provides special functions to hypothetically tests the update literals (rule heads) against the constraints.

To illustrate the syntax and semantics of ECA-LP, consider an ECA rule which states that "*every 10 seconds it is checked (time) whether there is an incoming request by a customer to book a flight to a certain destination (event). Whenever this event is detected, a database look-up selects a list of all flights to this destination (condition) and tries to book the first flight (action). In case this action fails, the system will backtrack and book the next flight in the list otherwise it succeeds (post-condition cut) sending a "flight booked" notification If no flight can be found to this destination, i.e. the condition fails, the else action is triggered, sending a "booked up" notification back to the customer.*" This is formalized in an ECA-LP as follows:

```
eca( every10Sec(), detect(request(Customer, Destination),T), find(Destination, Flight), book(Customer, Flight), !, notify(Customer, bookedUp(Destination) ).
```

% time derivation rule

```
every10Sec() :- sysTime(T), interval( timespan(0,0,0,10),T).

% event derivation rule
detect(request(Customer, FlightDestination),T):-
         occurs(request(Customer,FlightDestination),T),
         consume(request(Customer,FlightDestination)).

% condition derivation rule
find(Destination,Flight) :- on_exception(java.sql.SQLException,on_db_exception()),
dbopen("flights",DB), sql_select(DB,"flights", [flight, Flight], [where,
dest=Destination"]).

% action derivation rule
book(Cust, Flight) :-
         flight.BookingSystem.book(Flight, Cust),
         notify(Cust,flightBooked(Flight)).

% alternative action derivation rule
notify(Customer, Message):- sendMessage(Customer, Message).
```

The example includes possible backtracking to different variable bindings. If the action succeeds further backtracking is prevented by the post-conditional cut "!". If no flight can be found for the customer request, the else action is executed which notifies the customer about this. The condition derivation rule accesses an external data source via an SQL query.

### 3. The interval-based Event Calculus Event / Action Algebra.

Events resp. actions in reactive ECA rules are typically not atomic but are complex consisting of several atomic events or actions which must occur / be performed in the defined order and quantity in order to detect **complex events** resp. execute **complex actions**, e.g. an ordered sequence of events or actions. This topic has been extensively studied in the context of active databases and **event algebras**, which provide the operators to define complex event types. As we have pointed out in [2] typical event algebras of active database systems for complex event definitions such as Snoop [17] show inconsistencies and irregularities in their operators. For instance, consider the sequence *B;(A;C)* in Snoop. The complex event is detected if *A* occurs first, and then *B* followed by *C*, i.e. an event instance sequence (EIS) *EIS={a, b, c}* will lead to the detection of the complex event, because the complex event *(A;C)* is detected with associated detection time of the terminating event *c* and accordingly the event *b* occurs before the detected complex event *(A;C)*. However, this is not the intended semantics: Only the sequence *EIS={b, a, c}* should lead to the detection of the complex event defined by *B;(A;C)*. In the semantics of Snoop, which uses the detection time of the terminating event as occurrence time of a complex event, both complex event definitions *B;(A;C)* and *A(B;C)* are equal. This problem arises from the fact that the events, in the active database sense, are simply detected and treated as if they occur at an atomic instant, in contrast to the durative complex events, in the KR event/action logics sense, which occur over an extended interval. To overcome such unintended semantics and provide verifiable and traceable complex event computations (resp. complex actions) we have implemented a novel interval-based Event Calculus (EC) variant in the ContractLog KR and refined the typical event algebra operators based on it.

In the **interval-based Event Calculus** [1, 2, 4] all events are regarded to occur in an time interval, i.e. an event interval *[e1,e2]* occurs during the time interval *[t1,t2]* where *t1* is the occurrence time of *e1* and *t2* is the occurrence time of *e2*. In particular, an atomic event occurs in the interval *[t,t]*, where *t* is the occurrence time of the atomic event. The basic holdsAt axiom used in the EC [10] for temporal reasoning about fluents is redefined to the axiom *holdsInterval([E1,E2],[T1,T2])* to capture the semantics of event intervals which hold between a time interval:

```
holdsInterval([E1,E2],[T11,T22]):- event([E1],[T11,T12]), event([E2],[T21,T22]),
                [T11,T12]<=[T21,T22], not(broken(T12,[E1,E2],T21)).
```

The event function *event([Event],[Interval])* is a meta function to translate instantaneous event occurrences into interval-based events: event([E],[T,T]) :- occurs(E,T). It is also used in the event algebra meta program to compute complex events from occurred raw events according to their event type definitions. The *broken* function tests whether the event interval is not broken between the the initiator event and the terminator event by any other explicitly specified terminating event:

```
broken(T1,Interval,T2):-
        terminates(Terminator,Interval,[T1,T2]), event([Terminator],[T11,T12]), T1<T11,
T12<T2.
```

Example
```
occurs(a,datetime(2005,1,1,0,0,1)).
occurs(b,datetime(2005,1,1,0,0,10)).
Query: holdsInterval([a,b],Interval)?
Result: Interval = [datetime(2005,1,1,0,0,1), datetime(2005,1,1,0,0,10)]
```

Based on this interval-based event logics formalism, we now redefine the typical (SNOOP) event algebra operators and treat complex events resp. actions as occurring over an interval rather than in terms of their instantaneous detection times. In short, the basic idea is to split the occurrence interval of a complex event into smaller intervals in which all required component events occur, which leads to the definition of event type patterns in terms of interval-based event detection conditions, e.g. the sequence operator (;) is formalized as follows:

(A;B;C) ≡ detect(e,[T1,T3]) :- holdsInterval([a,b],[T1,T2],[a,b,c]), holdsInterval([b,c],[T2,T3],[a,b,c]), [T1,T2]<=[T2,T3].

In order to reuse detected complex events in rules, e.g. in ECA rules or other complex events, they need to be remembered until they are consumed, i.e. the contributing component events of a detected complex event should be consumed after detection of a complex event. This can be achieved via the previously described ID-based update primitives, which allow adding or removing knowledge from the KB. We use these update primitives to add detected event occurrences as new transient facts to the KB and consume events which have contributed to the detection of the complex event.

Example
```
detect(e,T):- event(sequence(a,b),T), % detection condition for the event e
         update(eis(e), "occurs(e,_0).", [T]), % add e with key eis(e)
         consume(eis(a)), consume(eis(b)). % consume all a and b events
```

If the detection conditions for the complex event *e* are fulfilled, the occurrence of the detected event *e* is added to the KB with the key *eis(e) (eis = event instance sequence)*. Then all events that belong to the type specific event instance sequences of type *a* and type *b* are consumed using their ids *eis(a)* resp. *eis(b)*. Different consumption policies are supported such as "remove all events which belong to a particular type specific *eis*" or "remove the first resp. the last event in the *eis*". If no consume predicate is specified in a detection rule, the events are reused in the detection of other complex events. For the reason of space we have only discusses processing of complex events, but the logical connectives defined by the interval-based event algebra are equally applicable on the definition of complex actions, e.g. to define a sequence of action executions (a sequence of transitions), with a declarative semantics for possibly required rollbacks, as defined above. For more information on ECA-LP see [1, 2].

### 4. ECA-RuleML: An ECA Rule Markup Language based on RuleML

Based on ECA-LP we now define **ECA-RuleML** [2] (in EBNF syntax), an **ECA rule markup language** extending RuleML. ECA-RuleML has been developed as a sublanguage of the Rule Based Service Level Agreement language (RBSLA - [18]). We follow the design principle of RuleML and define the new ECA-RuleML constructs within separated modules which are added to RuleML on top of the *hornlog layer* (extended with Naf and Neg) of RuleML as additional ECA-RuleML layers, i.e. ECA-RuleML adds additional expressiveness and modeling power to RuleML for the XML-based serialization of reactive rules (ECA rules), action language constructs, event notifications, event logics (Event Calculus) and event algebras (complex event/action pattern definitions).

hornlog2rbsla
*Naf ::= [oid,] weak | Atom | Cterm*
*Neg ::= [oid,] strong | Atom | Equal | Cterm*

Procedural Attachment
*Cterm ::= [oid,] op | Ctor | Attachment, {slot,} [resl,] {arg | Ind | Data |*
*Skolem | Var | Reify | Cterm | Plex}, [repo], {slot}, [resl]*
*Attachment ::= [oid,] Ind | Var | Cterm, Ind*

ID based update constructs
*Assert ::= content | And*
*Retract ::= oid*

ECA Rule Layer
*ECA ::= [oid,] [time,] [event,] [condition,] action [,postcondition] [,else]*
*time ::= Naf | Neg | Cterm*
*event ::= Naf | Neg | Cterm | Sequence | Or | Xor | Conjunction | Concurrent | Not | Any | Aperiodic*
*condition ::= Naf | Neg | Cterm*
*action ::= Cterm | Assert | Retract | Sequence | Or | Xor | Conjunction | Concurrent | Not | Any | Aperiodic*
*postcondition ::= Naf | Neg | Cterm*
*else ::= Cterm | Assert | Retract | Sequence | Or | Xor | Conjunction | Concurrent | Not | Any | Aperiodic*

Event Calculus Layer
*fluent ::= Ind | Var | Cterm*
*parameter ::= Ind | Var | Cterm*
*interval ::= Interval | Plex | Var*
*Happens ::= [oid,] event | Ind | Var | Cterm , time | Ind | Var | Cterm*
*Planned ::= [oid,] event | Ind | Var | Cterm, time | Ind | Var | Cterm*
*Occurs ::= [oid,], event | Ind | Var | Cterm, interval | Interval | Plex | Var*
*Initially ::= [oid,] fluent | Ind | Var | Cterm*
*Initiates ::= [oid,] event | Ind | Var | Cterm, fluent | Ind | Var | Cterm, time | Ind | Var | Cterm*
*Terminates ::= [oid,] event | Ind | Var | Cterm, fluent | Ind | Var | Cterm |*
*interval | Interval, time | Ind | Var | Cterm | interval | Interval*
*HoldsAt ::= [oid,] fluent | Ind | Var | Cterm , time | Ind | Var | Cterm*
*ValueAt ::= [oid,] parameter | Ind | Var | Cterm, time | Ind | Var | Cterm, Ind |*
*Var | Cterm | Data*
*HoldsInterval ::= [oid,] interval | Interval | Plex | operator | Sequence | Or |*
*Xor | Conjunction | Concurrent | Not | Any | Aperiodic | Periodic |*
*Cterm , interval | Interval | Plex | Var*
*Interval::= [oid,] event | time | Ind | Var | Cterm | operator | Sequence | Or |*
*Xor | Conjunction | Concurrent | Not | Any | Aperiodic | Periodic , event|*
*time | Ind | Var | Cterm | operator | Sequence | Or | Xor | Conjunction |*
*Concurrent | Not | Any | Aperiodic | Periodic*

Event Algebra Operators (defined in ECA Rule layer)
*operator ::= Sequence | Or | Xor | Conjunction | Concurrent | Not | Any | Aperiodic |*
*Periodic | Cterm*
*Sequence ::= [oid,] {event |action | Ind | Var | Cterm | operator | Sequence | Or | Xor*
*| Conjunction | Concurrent | Not | Any | Aperiodic | Periodic }*
*Or ::= [oid,] {event | action | Ind | Var | Cterm | operator | Sequence | Or | Xor*
*| Conjunction | Concurrent | Not | Any | Aperiodic | Periodic }*
*Conjunction ::= [oid,] {event | action | Ind | Var | Cterm | operator | Sequence | Or | Xor*
*| Conjunction | Concurrent | Not | Any | Aperiodic | Periodic }*
*Xor ::= [oid,] {event | action | Ind | Var | Cterm | operator | Sequence | Or | Xor*
*| Conjunction | Concurrent | Not | Any | Aperiodic | Periodic }*
*Concurrent:= [oid,] {event | action | Ind | Var | Cterm | operator | Sequence | Or |*
*Xor | Conjunction | Concurrent | Not | Any | Aperiodic | Periodic }*
*Not::= [oid,] event | action | Ind | Var | Cterm | operator | Sequence | Or | Xor | Conjunction |*
*Concurrent | Not | Any | Aperiodic | Periodic , interval | Interval | Plex | Var*
*Any ::= [oid,] Ind | Data | Var , event | action | Ind | Var | Cterm | operator | Sequence*
*| Or | Xor | Conjunction | Concurrent | Not | Any | Aperiodic | Periodic*
*Aperiodic ::= [oid,] event | action | Ind | Var | Cterm | operator | Sequence | Or | Xor |*
*Conjunction | Concurrent | Not | Any | Aperiodic | Periodic, interval |*
*Interval | Plex | Var*
*Periodic ::= [oid,] time | Ind | Var | Cterm, interval | Interval | Plex | Var*

ECA-RuleML is not intended to be executed directly, but transformed into the target execution language of an underlying rule-based systems (e.g. ECA-LP) and then executed there. We have implemented XSLT style sheets[6] which transform RuleML and ECA-RuleML into ECA-LPs resp. ContractLog/Prova scripts.

---

[6] http://ibis.in.tum.de/staff/paschke/rbsla/index.htm

## 5. Conclusion

In this paper we have described ECA-LP and ECA-RuleML a **homogenous reactive logic programming update language**. It uses logic programming techniques and homogenously combines derivation rules and reactive ECA rules together with several other (non-monotonic) formalisms such as integrity constraints or an interval-based Event Calculus formalization which used for temporal reasoning over the effects of events and actions. A generalized ECA interpreter which implements the operational semantics of the ECA paradigm is provided as an add-on to arbitrary LP systems. It supports parallel ECA rule execution, transactional (bulk) ID-based updates and complex event processing based on transient and non-transient events as well as temporal event reasoning by KR-based event / action logics. Due to the homogeneous LP semantics ECA-LP can be easily combined with other logical formalisms. This clearly leads to higher levels of expressiveness than traditional ECA systems provide. The approach is in particular suitable in domains where the ability to reason formally about the event/action logic is crucial for describing reliable and predictable real-world decision logic and trigger rule based reactions in response to events and actions.